# A Dynamic Linear Bias Incorporation Scheme for Nonnegative Latent Factor Analysis

Yurong Zhong, Zhe Xie, Weiling Li, and Xin Luo

**Abstract**—High-Dimensional and Incomplete (HDI) data is commonly encountered in big data-related applications like social network services systems, which are concerning the limited interactions among numerous nodes. Knowledge acquisition from HDI data is a vital issue in the domain of data science due to their embedded rich patterns like node behaviors, where the fundamental task is to perform HDI data representation learning. Nonnegative Latent Factor Analysis (NLFA) models have proven to possess the superiority to address this issue, where a linear bias incorporation (LBI) scheme is important in present the training overshooting and fluctuation, as well as preventing the model from premature convergence. However, existing LBI schemes are all statistic ones where the linear biases are fixed, which significantly restricts the scalability of the resultant NLFA model and results in loss of representation learning ability to HDI data. Motivated by the above discoveries, this paper innovatively presents the dynamic linear bias incorporation (DLBI) scheme. It firstly extends the linear bias vectors into matrices, and then builds a binary weight matrix to switch the active/inactive states of the linear biases. The weight matrix's each entry switches between the binary states dynamically corresponding to the linear bias value variation, thereby establishing the dynamic linear biases for an NLFA model. Empirical studies on three HDI datasets from real applications demonstrate that the proposed DLBI-based NLFA model obtains higher representation accuracy several than state-of-the-art models do, as well as highly-competitive computational efficiency.

———————————— ◆ ————————————

## I. INTRODUCTION

High-Dimensional and Incomplete (HDI) data are prevalent data-related applications like recommender systems [23, 29-32] and Quality-of-Service (QoS) predictor in web service [19, 22, 25]. Note that HDI data is typically characterized by:
1. Its entities set is large;
2. Its most data entries are missing; and
3. Its data are commonly nonnegative like recommender system's ratings [29-32].

Despite its incompleteness, valuable knowledge like user-item preferences [29-32] can be extracted. Hence, an analysis model should well represent HDI data for extracting such knowledge.

For effectively represent HDI data, recent studies' proposed models can mainly be split into three general categories: 1) Graph Convolutional Network (GCN) models [2, 4]. For example, Wang *et al*.'s NGCF [4] and He *et al*.'s LightGCN [2] are able to capture the node information from the graph and extract nonlinear features from HDI data; 2) Nonnegative Matrix Factorization (NMF) models [5, 12]. Yang *et al*.'s B-NMF [5] and Cai *et al*.'s ANMF [12] factorizes nonnegative HDI data into nonnegative Latent Factor (LF) matrices in a low-rank way, which is iteratively solved by a Nonnegative and Multiplicative Update (NMU) algorithm for achieving valuable knowledge. In spite of their effectiveness and efficiency, they do not consider the incompleteness of HDI data; 3) Nonnegative Latent Factor Analysis (NLFA) models [6]. Their learning objective is defined on observed interactions only. Then, a Single LF-dependent Nonnegative and Multiplicative Update (SLF-NMU) algorithm is designed for efficiently solving such a learning objective.

Furthermore, a linear bias incorporation (LBI) scheme has proven to be effective in preventing NLFA from the training overshooting and fluctuation for well convergence [48, 49], such as Luo *et al*.'s BNLFA [48] and Chen et al.'s EBNL [49]. Note that these LBI schemes are all statistic ones where the linear biases are fixed, which significantly restricts the scalability of the resultant NLFA model and results in loss of representation learning ability to HDI data. Aiming at addressing above critical issues, this paper proposes a dynamic linear bias incorporation (DLBI) scheme. The main contribution of this paper includes:
1. The DLBI scheme for NLFA. It firstly extends the linear bias vectors into matrices, and then builds a binary weight matrix whose each entry switches between the binary states dynamically according to variation of the linear bias value, thereby achieving a DLBI-based NLFA (DNLFA) model;
2. It performs empirical studies on three real-world nonnegative HDI data to demonstrate that a DNLFA model achieves higher representation accuracy than state-of-the-art models do, as well as promising computational efficiency.

Section II introduces the preliminaries. Section III presents a PSNL model. Section IV gives the experimental results. Finally, Section V concludes this paper.

————————————————————————————

- ***Corresponding author: X. Luo.***
- *Y. Zhong, Z. Xie, W. Li and X. Luo are with the School of Computer Science and Technology, Dongguan University of Technology, Dongguan, Guangdong 523808, China (e-mail: zhongyurong91@gmail.com, gxyz4419@gmail.com, weilinglicq@outlook.com, luoxin21@gmail.com).*

## II. PRELIMINARIES

*A. Problem Formulation*

HDI data filled with nonnegative values is as defined next [6, 14, 20, 23].

**Definition 1.** Given $M$ and $N$, $R^{|M|\times|N|}$ describes user-item interactions among them and these interactions are weighted. Given known set $\Lambda$ and unknown set $\Gamma$ for $R$, $R$ is nonnegative HDI data if $|\Lambda|\ll|\Gamma|$.

**Definition 2.** Given $R$ and $\Lambda$, an NLFA model [6] usually relies on $\Lambda$ to seek for rank-$D$ approximation of $R$, i.e. $\hat{r}_{m,n}=\sum_{k=1}^{d_1}x_{m,k}y_{n,k}$, where $x_{m,k}\geq 0$ and $y_{n,k}\geq 0$. With commonly-used Euclidean distance [7-9, 38-40] and $L_2$-norm-based regularization scheme [11, 21, 24, 36, 37], the following objective function is defined as:

$$\varepsilon = \frac{1}{2}\sum_{r_{m,n}\in\Lambda}\left((r_{m,n}-\hat{r}_{m,n})^2+\lambda\sum_{k=1}^{d_1}\left((x_{m,k})^2+(y_{n,k})^2\right)\right),$$

$$s.t. \quad \forall m\in\{1,2,\ldots,|M|\}, n\in\{1,2,\ldots,|N|\}, k\in\{1,2,\ldots,d_1\}: x_{m,k}\geq 0, y_{n,k}\geq 0,$$

(1)

where the $L_2$-norm-based regularization coefficient $\lambda$ is positive.

**Definition 3.** For implementing LBI in NLFA, two linear bias vectors $C^{|M|}$ and $F^{|N|}$ are assigned for NLFA's $X^{|M|\times d1}$ and $Y^{|N|\times d1}$, respectively. Hence, $\forall c_m\in C$ and $f_n\in F$, (1) can be extended as follows:

$$\varepsilon = \frac{1}{2}\sum_{r_{m,n}\in\Lambda}\left(\left(r_{m,n}-\sum_{k=1}^{d_1}x_{m,k}y_{n,k}-c_m-f_n\right)^2+\lambda\left((c_m)^2+(f_n)^2+\sum_{k=1}^{d_1}\left((x_{m,k})^2+(y_{n,k})^2\right)\right)\right),$$

$$s.t. \quad \forall m\in\{1,2,\ldots,|M|\}, n\in\{1,2,\ldots,|N|\}, k\in\{1,2,\ldots,d_1\}:$$

$$x_{m,k}\geq 0, y_{n,k}\geq 0, c_m\geq 0, f_n\geq 0.$$

(2)

Hence, with (2), a BNLFA model is achieved.

## III. A DNLFA MODEL

First of all, we extend linear bias vectors $C$ and $F$ adopted in (2) into linear bias matrices $G^{|M|\times d2}$ and $H^{|N|\times d2}$. Then, two weighted matrices $I^{|M|\times d2}$ and $J^{|N|\times d2}$ are introduced for $G$ and $H$, respectively. Note that $I$ and $J$ are manipulated with the following principle:

1. $\forall\ i_{m,k}\in I=1$ and $j_{n,k}\in J=1$; and
2. if $\forall\ g_{m,k}\geq e$, $i_{m,k}$ keeps unchanged, otherwise $i_{m,k}=0$, where $m\in\{1,2,\ldots,|M|\}$, $k\in\{1,2,\ldots,d_2\}$ and $e>0$. Note that similar designs are also applied to $J$.

With such a design, DLBI is compatible for an NLFA model. Hence, with DLBI, (1) can be reformulated as:

$$\varepsilon = \frac{1}{2}\sum_{r_{m,n}\in\Lambda}\left(\left(r_{m,n}-\sum_{k_1=1}^{d_1}x_{m,k_1}y_{n,k_1}-\sum_{k_2=1}^{d_2}\left(i_{m,k_2}g_{m,k_2}+j_{n,k_2}h_{n,k_2}\right)\right)^2+\lambda\left(\sum_{k_1=1}^{d_1}\left((x_{m,k_1})^2+(y_{n,k_1})^2\right)+\sum_{k_2=1}^{d_2}\left((i_{m,k_2}g_{m,k_2})^2+(j_{n,k_2}h_{n,k_2})^2\right)\right)\right),$$

$$s.t. \quad \forall m\in\{1,2,\ldots,|M|\}, n\in\{1,2,\ldots,|N|\}, k_1\in\{1,2,\ldots,d_1\}, k_2\in\{1,2,\ldots,d_2\}:$$

$$x_{m,k_1}\geq 0, y_{n,k_1}\geq 0, g_{m,k_2}\geq 0, h_{n,k_2}\geq 0, i_{m,k_2}=\{0,1\}, j_{n,k_2}=\{0,1\}.$$

(3)

To efficiently solve DNLFA's learning objective, an SLF-NMU algorithm is adopted [6, 10, 18, 33]. First of all, an additive gradient descent (AGD) algorithm is applied to (3), yielding the following learning rules:

$$\underset{X,Y,G,H}{\arg\min}\ \varepsilon\ \overset{AGD}{\Rightarrow}$$

$$\forall m\in\{1,2,\ldots,|M|\}, n\in\{1,2,\ldots,|N|\}, k_1\in\{1,2,\ldots,d_1\}, k_2\in\{1,2,\ldots,d_2\}:$$

$$\begin{cases} x_{m,k_1}\leftarrow x_{m,k_1}+\eta_{m,k_1}\sum_{n\in\Lambda(m)}\left(y_{n,k_1}(r_{m,n}-\hat{r}_{m,n})-\lambda x_{m,k_1}\right), \\ y_{n,k_1}\leftarrow y_{n,k_1}+\eta_{n,k_1}\sum_{m\in\Lambda(n)}\left(x_{m,k_1}(r_{m,n}-\hat{r}_{m,n})-\lambda y_{n,k_1}\right), \\ g_{m,k_2}\leftarrow g_{m,k_2}+\eta_{m,k_2}\sum_{n\in\Lambda(m)}\left(i_{m,k_2}(r_{m,n}-\hat{r}_{m,n})-\lambda g_{m,k_2}\right), \\ h_{n,k_2}\leftarrow h_{n,k_2}+\eta_{n,k_2}\sum_{m\in\Lambda(n)}\left(j_{n,k_2}(r_{m,n}-\hat{r}_{m,n})-\lambda h_{n,k_2}\right). \end{cases}$$

(4)

For canceling these negative terms, by manipulating $\eta_{m,k1}$, $\eta_{n,k1}$, $\eta_{m,k2}$ and $\eta_{n,k2}$, (4) can be reformulated as:

$$x_{m,k_1}\leftarrow x_{m,k_1}\left(\sum_{n\in\Lambda(m)}(r_{m,n}y_{n,k_1})\Big/\sum_{n\in\Lambda(m)}(\hat{r}_{m,n}y_{n,k_1}+\lambda x_{m,k_1})\right),$$ (5a)

$$y_{n,k_1}\leftarrow y_{n,k_1}\left(\sum_{m\in\Lambda(n)}(r_{m,n}x_{m,k_1})\Big/\sum_{m\in\Lambda(n)}(\hat{r}_{m,n}x_{m,k_1}+\lambda y_{n,k_1})\right),$$ (5b)

$$g_{m,k_2} \leftarrow g_{m,k_2} \left( \sum_{n \in \Lambda(m)} r_{m,n} i_{m,k_2} \Big/ \sum_{n \in \Lambda(m)} \left( \hat{r}_{m,n} i_{m,k_2} + \lambda g_{m,k_2} \right) \right), \quad (5c)$$

$$h_{n,k_2} \leftarrow h_{n,k_2} \left( \sum_{m \in \Lambda(n)} r_{m,n} j_{n,k_2} \Big/ \sum_{m \in \Lambda(n)} \left( \hat{r}_{m,n} j_{n,k_2} + \lambda h_{n,k_2} \right) \right). \quad (5d)$$

After updating $g_{m,k2}$ and $h_{n,k2}$ at the $f$-th iteration, we update $i_{m,k2}$ and $j_{n,k2}$ at the current iteration with the following learning rules:

$$\begin{cases} i^f_{m,k_2} = 0 & if \ g^f_{m,k_2} < e, \\ j^f_{n,k_2} = 0 & if \ h^f_{n,k_2} < e. \end{cases} \quad (6)$$

## IV. EXPERIMENTAL RESULTS AND ANALYSIS

### A. General Settings

**Evaluation Protocol.** In real-world applications, decomposing HDI data into LFs is critical for predicting missing values and identifying potential connections between entities [26-28, 34, 35]. As a result, this technique is often employed as an evaluation protocol for assessing the performance of related models.

**Evaluation Metrics.** The accuracy of a test model for missing data predictions can be measured by the root mean square error (RMSE) [14-17, 47]:

$$RMSE = \sqrt{\left( \sum_{r_{m,n} \in \Gamma} (r_{m,n} - \hat{r}_{m,n})^2 \right) \Big/ |\Gamma|},$$

where $\Gamma$ denotes the validation set and is disjoint with the training set $\Lambda$. Note that low RMSE represents high prediction accuracy for missing data in $\Gamma$.

**Datasets.** Our experiments adopt HDI data in real big data-related applications, and their details are shown in Table I. In all experiments on each dataset, we randomly split its known set of entries $\Lambda$ into ten disjoint subsets for tenfold cross-validation. We adopt seven subsets as the training set, one subset as the validation set, and the remaining two subsets as the test set. This process is repeated ten times sequentially.

TABLE I. Details of Adopted Datasets.

| No. | Name | $|\Lambda|+|\Gamma|$ | $|U|$ | $|I|$ | Density | Source |
|---|---|---|---|---|---|---|
| D1 | EM | 2,811,718 | 61,265 | 1,623 | 2.83% | EachMovie |
| D2 | Flixter | 8,196,077 | 147,612 | 48,794 | 0.11% | [1] |
| D3 | Douban | 16,830,839 | 129,490 | 58,541 | 0.22% | [3] |

TABLE II. Details of Tested Models.

| No. | Name | Description |
|---|---|---|
| M1 | NLFA | A standard NLFA model [6]. |
| M2 | NIR | A recent NLFA model [46]. |
| M3 | BNLFA | A LBI-based NLFA model with linear bias vectors [48]. |
| M4 | EBNL | A LBI-based NLFA model with linear bias matrices [49]. |
| M5 | LightGCN | A commonly-adopted Graph Convolutional Network model [2]. |
| M6 | PSNL | The proposed model in this paper. |

**Compared Models.** Our experiments involve six models, and their details are presented in Table II. To achieve its objective results, we used the following settings:

(1) The proposed DNLFA's $d_1$ and $d_2$ are set at 20 and 5, respectively. Considering compared model, their LF dimension is set at 20 uniformly;
(2) For each model on each data set, the results generated from 10 different random initial values are recorded to calculate the average RMSE and convergence time for eliminating the effect of initial assumptions [13, 41-45].
(3) The training process of the test model is terminated when: 1) the iteration count reaches a preset threshold, which is 1000; 2) The difference between the two consecutive iterations of the generated RMSE is less than $10^{-5}$.

### B. Comparison against State-of-the-art Models

Table III and IV summarize RMSE and total time cost of M1-6 on D1-3, respectively. From these results, we have the following findings:

(1) **DNLFA achieves significantly higher representation accuracy than state-of-the-art models do.** For example, as shown in Table III, RMSE of M6 is 0.2339 on D1, which is about 0.55%, 2.7%, 0.26%, 1.27% and 3.86% lower than M1's 0.2352, M2's 0.2404, M3's 0.2345, M4's 0.2369 and M5's 0.2433, respectively. Similar results can be found on D2 and D3. Hence, DNLFA achieves significantly higher representation accuracy than state-of-the-art models do.
(2) **DNLFA's computational efficiency is highly-competitive.** DNLFA's total time cost is the least on D2, and it has the third least total time cost on D1 and D3, as recorded in Table IV.

TABLE III. RMSE of M1-6 on D1-3.

| No. | M1 | M2 | M3 | M4 | M5 | M6 |
|---|---|---|---|---|---|---|
| D1 | 0.2352 ±9.3E-5 | 0.2404 ±9.0E-3 | 0.2345 ±6.5E-4 | 0.2369 ±6.8E-4 | 0.2433 ±3.7E-4 | **0.2339 ±5.4E-5** |
| D2 | 0.9619 ±3.5E-4 | 0.9592 ±7.4E-4 | 0.9502 ±4.9E-4 | 0.9422 ±1.1E-5 | 1.0216 ±9.5E-4 | **0.9268 ±5.7E-5** |
| D3 | 0.7285 ±2.8E-4 | 0.7284 ±2.6E-4 | 0.7256 ±8.1E-4 | 0.7286 ±1.1E-5 | 0.7681 ±5.1E-4 | **0.7207 ±2.5E-4** |

TABLE IV. Total Time Cost of M1-6 on D1-3 (Seconds).

| No. | M1 | M2 | M3 | M4 | M5 | M6 |
|---|---|---|---|---|---|---|
| D1 | 127 ±19.65 | **96 ±9.87** | 223 ±24.16 | 314 ±46.92 | 2,795 ±218.37 | 171 ±14.19 |
| D2 | 840 ±73.65 | 1,624 ±165.64 | 123 ±15.12 | 293 ±26.37 | 17,506 ±1529.95 | **89 ±9.44** |
| D3 | 1,890 ±211.32 | 1,941 ±182.61 | 840 ±78.36 | **195 ±17.45** | 32,857 ±2295.36 | 884 ±101.06 |

## V. CONCLUSIONS

A DNLFA model has shown great potential in predicting missing data of nonnegative HDI data in real applications. Its highly accurate representation boosts its practicability. Note that a DNLFA model's performance depends on the choice of its hyper-parameters, i.e., *λ* and *e*. Hence, it is highly necessary to investigate computing intelligence approaches for making them self-adaptive in the future.